\documentclass[letterpaper, 10 pt, conference]{ieeeconf} 
\IEEEoverridecommandlockouts                              % This command is only needed if 
\usepackage{graphicx}
\usepackage{amsmath}
\usepackage{algorithm}
\usepackage{algpseudocode}
\usepackage{xcolor}
\usepackage{soul}
\usepackage{stfloats}
\usepackage{amsfonts} 
\usepackage{amssymb}
\usepackage{tabularx}
\usepackage{booktabs}
\usepackage{comment}
\usepackage{subcaption}
\usepackage{xspace}
\usepackage{cite}

\usepackage{xcolor}
\usepackage{comment}
\usepackage{hyperref}

\usepackage{indentfirst}
\usepackage{todonotes}
\usepackage{tkz-graph}
\usetikzlibrary{automata,positioning}
\usepackage{tikz}
\usetikzlibrary{positioning}
\usepackage[nolist,nohyperlinks]{acronym}

\title{\LARGE \bf
Listen to Your Map: \\ 
An Online Representation for Spatial Sonification}

\author{Lan Wu$^{1}$, Craig Jin$^{2}$, Monisha Mushtary Uttsha$^{1}$ and Teresa Vidal-Calleja$^{1}$%
\thanks{$^{1}$Authors are with the Robotics Institute, University of Technology Sydney, Ultimo, NSW 2007, Australia.}
\thanks{$^{2}$Author is with the Computing and Audio Research Laboratory, University of Sydney, Camperdown, NSW 2050, Australia.}
\thanks{This work was supported by ARIA Research and the Australian Government via the Department of Industry, Science, and Resources CRC-P program (CRCPXI000007).} \thanks{\tt\footnotesize Lan.Wu-2@uts.edu.au}
}

\begin{document}

\maketitle
\thispagestyle{empty}
\pagestyle{empty}

%%%%%%%%%%%%%%%%%%%%%%%%%%%%%%%%%%%%%%%%%%%%%%%%%%%%%%%%%%%%%%%%%%%%%%%%%%%%%%%%
\begin{abstract}

%We aim to build a representation to connect the dots between collecting information via robotic vision and then sonifying the spatial scene properly.
%TVC thanks copilot :-D Lan: haha
Robotic perception is becoming a key technology for navigation aids, especially helping individuals with visual impairments through spatial sonification. This paper introduces a mapping representation that accurately captures scene geometry for sonification, turning physical spaces into auditory experiences. Using depth sensors, we encode an incrementally built 3D scene into a compact 360-degree representation with angular and distance information, aligning this way with human auditory spatial perception. The proposed framework performs localisation and mapping via VDB-Gaussian Process Distance Fields for efficient online scene reconstruction. The key aspect is a sensor-centric structure that maintains either a 2D-circular or 3D-cylindrical raster-based projection. This spatial representation is then converted into binaural auditory signals using simple pre-recorded responses from a representative room. Quantitative and qualitative evaluations show improvements in accuracy, coverage, timing and suitability for sonification compared to other approaches, with effective handling of dynamic objects as well. An accompanying video demonstrates spatial sonification in room-like environments.\footnote[1]{\tt \url{https://tinyurl.com/ListenToYourMap}}

\end{abstract}

\begin{figure}[t]
  \centering
  \resizebox{0.9\linewidth}{!}{
  \subfloat[]{\includegraphics[height=5cm]{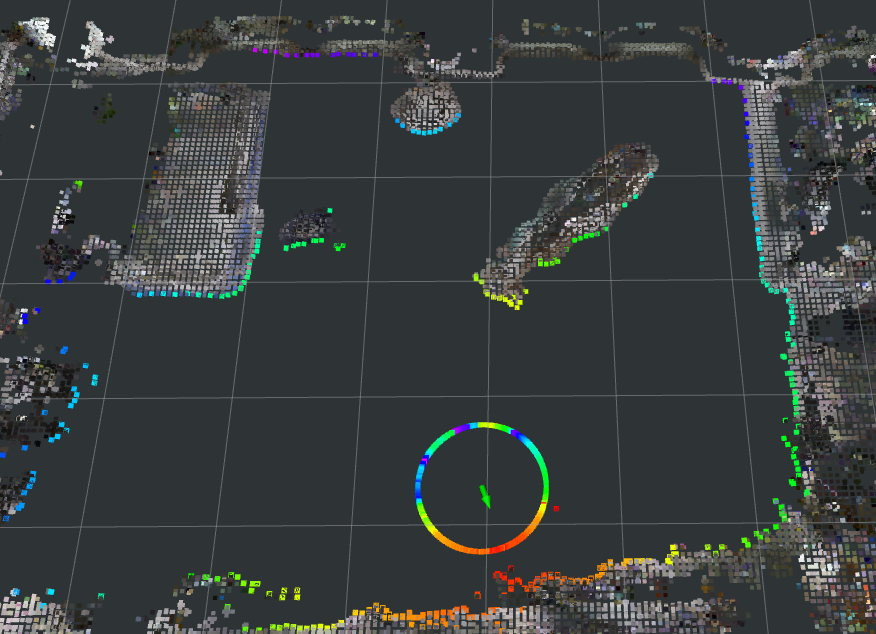}}}
  \resizebox{0.9\linewidth}{!}{
  \subfloat[]{\includegraphics[height=5cm]{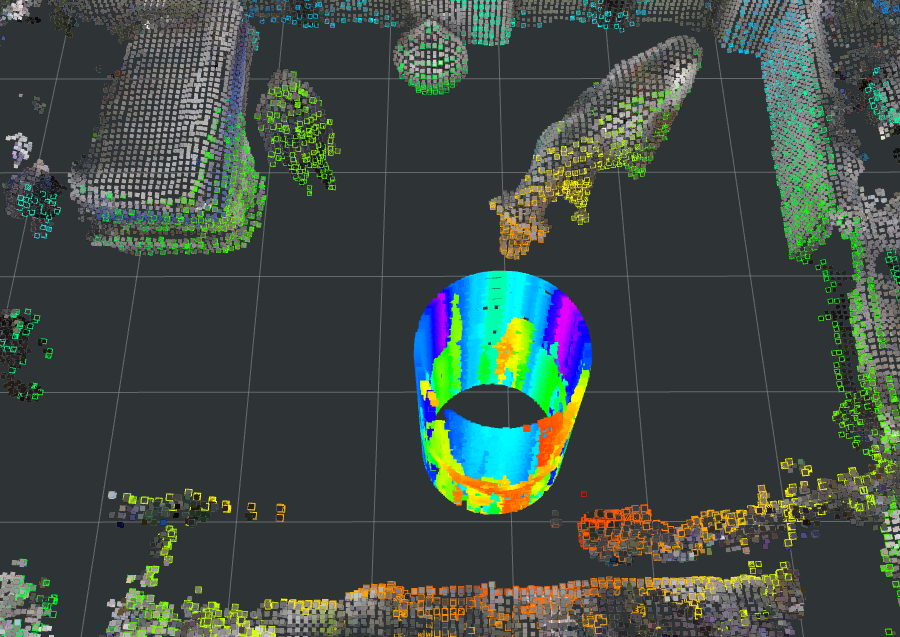}}
  }
  \caption{Our sensor-centric representation for spatial sonification. a) 2D circular representation and b) 3D cylindrical representation for \emph{the Cow and Lady} dataset. The circle is coloured by the distance to the obstacle along each azimuthal angle. We also show the selected close points on the surface. Similarly, for the cylinder, we illustrate the distance up to a certain height.}
  \label{teaser}
  %\vspace{-3ex}
\end{figure}

%%%%%%%%%%%%%%%%%%%%%%%%%%%%%%%%%%%%%%%%%%%%%%%%%%%%%%%%%%%%%%%%%%%%%%%%%%%%%%%%
%Augmented Reality in Audio: Spatial computing and AI converge to empower agency and independence for people living with vision disability.
\section{Introduction}\label{sec:introduction}

% Motivation
% - Sensory augmentation for visually impaired people
% - 
% \\
Robotic perception has the potential to interpret the environment on behalf of humans. Over the past two decades, robotic sensors such as cameras or LiDARs have acted as the eyes of robots, becoming essential tools for gathering information on unknown environments. For individuals living with visual impairments, auditory perception is crucial. This need could be fulfilled by transforming the information accumulated by the robotic perception algorithms via sonification. 
One of the key steps in representing an environment for sonification is to convert spatial data into an organised format that can be understood through sound. This process facilitates the transformation of physical spaces into auditory experiences, allowing users to understand and navigate through sound. The primary goal of this work is to build a robust framework that can precisely represent an environment's spatial properties such that it can serve as a basis for sonification. 

The ability to take a light-weight sensor information and incrementally build and maintain a compact representation will enable real-time auditory feedback to support spatial awareness and environment interaction, especially for visually impaired individuals. As we want our representation to cater towards sonification for humans, we take motivation from how human beings perceive an environment through sound. Thus, we transform an incrementally built 3D scene captured with depth data (from an RGB-D camera for instance) into a 360-degree projection of distance to the surface information. This approach naturally aligns with human auditory perception while simplifying information to focus on the most relevant spatial points. To achieve this, we propose a mapping representation, based on a sensor-centric organised structure that maintains 2D circular and/or 3D cylindrical rasterised representations (see Fig.~\ref{teaser}). These structures capture the geometry of the 3D environment and offer flexibility in how spatial locations are sonified by leveraging VDB-GPDF~\cite{wu2024vdb}, an online efficient mapping framework based on Gaussian Process (GP) distance and gradient fields and a fast-access VDB data structure (Volumetric, Dynamic grid that shares several characteristics with B+trees~\cite{museth2013vdb,museth2013openvdb}). In addition, a localisation framework based on VINS-RGBD~\cite{vins_rgbd} is combined with our mapping approach to provide a full online Simultaneous localisation and mapping (SLAM) solution with synchronised poses and raw dense depth data. %It provides online and synchronised poses and raw dense depth measurements for the mapping to incrementally build the representations. 
Moreover, by incorporating custom binaural room impulse responses (BRIRs), our online incrementally-built scene representation offers perceptually robust auditory augmentation of the mapped environment for both near-field and
far-field obstacles and unknown space.  %enabling a sonifiable online incrementally-built scene representation for indoor and outdoor scenarios. 
% Our probabilistic framework efficiently fuses surface points from depth-only information into the Euclidean distance field; it has the ability to fuse other surface properties, including intensity, colour, etc., and can reconstruct the scene mesh with informative textures. 

% The objective is to facilitate smooth integration with sonification algorithms by representing spatial data into a sonifiable format. 
%Moreover, we demonstrate two modes of operation, a {\em circular ranging} and a {\em circular ranging of objects} sonification modes, which sonify the distance to objects for various directions around the circle.
%We also introduce various filtering operations to extract information from the proposed 2D circular representation. 

We evaluate quantitatively the performance of the proposed mapping representation in terms of accuracy, coverage and timing, demonstrating the ability to convey the required spatial information better than naively using the current sensor information or a state-of-the-art Euclidean Distance Field. In addition, we show qualitative performance in the presence of dynamic objects and benchmarked with the information provided by distance field mapping. Finally, the submitted video showcases the spatial sonification for room-like environments.

\section{Literature Review}
 %%%%%%%representation focused paper
 Several studies have explored sonification from vision-based representations. While earlier works focused mostly on 2D vision-to-audio transformations \cite{ward2010visual},  more recent approaches utilise 3D vision-based sensors \cite{yang2018long,li2020wearable}. Although 3D representations generally offer improved accuracy for localisation tasks, 2D systems can achieve comparable performance for navigation with sonification \cite{commere2022sonified}. Some works have directly used 3D raw data for environment representation \cite{brock2013supporting,stoll2015navigating}. In \cite{brock2013supporting}, the authors presented a real-time auditory feedback system that assists blind users in obstacle navigation using a depth camera. The system processes depth data to detect obstacles and sonifies this information. However, using depth data directly can be complex for turning 3D information into clear auditory cues as the information over time can be inconsistent. To tackle this, we propose to reconstruct the environment with sequential depth data and encode it into a consistent 360-degree circular/cylindrical raster-based projection. 
 
 More recently, there has been a growing trend towards leveraging machine learning techniques, such as generative models, to enhance environment representation for spatial sonification. These methods allow for more adaptive and context-aware sonification systems, improving the overall user experience. In \cite{dagli2024see}, the authors introduced ``SEE-2-SOUND," a zero-shot framework. They generated spatial audio from visual inputs such as images, GIFs, and videos without pre-training.  %real-world audio, 
 A similar task was performed in \cite{su2024sonifyar}, with the additional use of a large language model to generate context-aware synchronised audio-visual content based on the event. While these systems offer adaptability, they often require substantial computational resources. %This introduces substantial latency in the sonification process. 
 In contrast, our method provides a lightweight, direct mapping representation of spatial data to auditory cues. This ensures real-time responsiveness while preserving essential spatial details and the ability to be deployed on modest embedded devices.
 % producing synchronized  by estimating 3D locations.
 
 % Horizontal and vertical positions are conveyed via stereo panning and pitch, respectively, while distance is encoded by volume.
 % Tested with both sighted and blind participants, the system demonstrated that users could learn to navigate effectively with minimal training.
 
 %%%%%Different sonification paper
 There are different techniques to sonify a virtual environment, Gao et al. showed quantitative evidence of how different sonification strategies can influence user perception and behaviour in~\cite{gao2022exploring}. They compared four elevation sonification mapping strategies and found that binary relative mapping performs better than the others, and in combination with azimuth sonification, it gives clear directional information generating more accurate and efficient task completion. Keeping this in mind, our proposed representation technique is designed to effectively capture both angular and distance information. Thus, ensuring optimal translation of spatial data into auditory cues for the sonification process.

 %%%%sonification and representation methods integrated
 %%
 %%for visually impaired
 There have been many works on sonification that are dedicated towards visually impaired persons\cite{schwartz2024echosee}. The type of sound used for sonification also varies depending on the task and the type of information it carries. While discrete sounds are good for simpler information and objects that are near the user, continuous sounds are more suited to deliver complex elements in the scene \cite{peetoom2015disability}. Schwartz et al. proposed a mobile application that can reconstruct and sonify a scene to guide users with visual impairment in real-time \cite{schwartz2024echosee}. The work in \cite{zhao2019multi} built a 3D sonification framework to sonify images from video sequences to send the user a predefined sonic sensorial description through bone-conduction headphones. They combined YOLOV3 \cite{redmon2018yolov3} for object detection with a probabilistic occupancy map from OctoMap~\cite{Octomap} to detect environmental anomalies. These works primarily focus on sonifying immediate surroundings or predefined objects with limited adaptability. They do not offer a full 360-degree spatial coverage, focusing mostly on near-field obstacles. Our proposed approach, in contrast, is flexible, robust and offers greater adaptability as it conveys only spatial information for near-field and far-field.\\
 % Most of these works focus on sonifying immediate surroundings, but \cite{gupta2024sonicvista} introduces a method that emphasizes enhancing auditory awareness of distant objects like buildings, birds, and trees. 

 % Some of these works use hardware setups with different wearable parts as part of the system.

\section{Preliminaries}
\subsection{Localisation}
We employ a modified VINS-RGBD~\cite{vins_rgbd} localisation framework to generate precise camera poses as inputs for the mapping. A key aspect of our approach is the improved synchronisation between the estimated poses and the raw dense point cloud data captured by the RGB-D camera. By refining the VINS framework, we ensure consistent spatial alignment, which enhances the accuracy and reliability of the mapping process, ultimately enabling more precise and detailed scene reconstruction.

\subsection{VDB-GPDF Mapping}
%Our representation is based on the VDB-GPDF mapping framework as proposed in \cite{wu2024vdb}. It couples the VDB grid structure with the Gaussian Process Distance Field to gradually construct and maintain a large-scale, dense distance field map of the environment. The depth sensor data at world frame $\mathcal{P}_{\{i\}}$   
%is voxelised to create a local VDB structure $\mathcal{P}_{v\{i\}}$. In every VDB leaf node, the local voxel centres act as training points for GPs. The collection of GPs across all leaf nodes forms the temporary latent Local GP Signed Distance Field (L-GPDF).

%For testing, a set of points $\mathcal{P}_{t\{i\}}$ is generated by ray-casting from the sensor origin through the voxels in the current frame $\mathcal{P}_{v\{i\}}$. Each test point $ \mathbf{x}^* \in \mathcal{P}_{t\{i\}}$ is used to query the L-GPDF for the distance field \( \hat{d}_t(\mathbf{x}^*) \), surface properties \( \hat{c}_t(\mathbf{x}^*) \), and variance estimates \( \hat{v}_t(\mathbf{x}^*) \). 
%The queried values from L-GPDF are then fused into a global VDB grid map, $ \mathcal{P}_{\{0,...,i\}}$, by updating the voxel distances using the weighted sum method. Following the fusion of the distances and surface properties, the marching cubes algorithm is used to reconstruct a dense surface by identifying the active voxels in the global VDB. The Global Gaussian Process Distance Field (G-GPDF) is made up of the zero-crossing points from active leaf nodes. At this stage, queries for the G-GPDF are computed by averaging the inferred distances from neighboring GP nodes.

Our representation is based on the VDB-GPDF mapping framework as proposed in \cite{wu2024vdb}. It couples the VDB grid structure with the Gaussian Process Distance Field~\cite{wu2021faithful,wu_log-gpis-mop_2023,legentil2023accurate} to gradually construct and maintain a large-scale, dense distance field map of the environment. The depth sensor data (from depth cameras or LiDARs) at the world coordinate frame    
is voxelised to create a local VDB structure with 3D points. In every VDB leaf node, the local voxel centres act as training points for GPs. The collection of GPs across all leaf nodes forms the temporary latent Local GP Signed Distance Field (L-GPDF).

For testing, a set of query points is generated by ray-casting from the sensor origin through the voxels in the current reference frame. Each test point is used to query the L-GPDF for the distance field and variance estimates. 
The queried values from L-GPDF are then fused into a global VDB grid map by updating the voxel distances 
using the weighted sum method. Following the fusion of the distances and surface properties, the marching cubes algorithm~\cite{marching} is used to reconstruct a dense surface by identifying the active voxels in the global VDB. The Global Gaussian Process Distance Field (G-GPDF) is made up of the zero-crossing points from active leaf nodes. At this stage, queries for the G-GPDF are computed by averaging the inferred distances from neighbouring GP nodes. For further details refer to~\cite{wu2024vdb}.

%\section{Sound Range Scanner}
\section{Circular/Cylindrical Rasterisation}
%\subsection{Overview}
We aim to incrementally build a scene representation that facilitates spatial sonification of an unknown environment. We propose a representation based on VDB-GPDF that is encoded to cater for the sonification requirements of future human sensory augmentation. %Given a 3D point cloud from a depth camera or a LiDAR in the sensor's local coordinate frame, the data is transformed into the world reference frame. The dense point cloud data is voxelised into a hierarchical VDB structure to manage it more effectively. A Local Gaussian Process Signed Distance Field (L-GPDF), which models surface properties and distance at the local level, is generated on each voxelised point cluster. Following their integration, these local fields form the Global GPDF (G-GPDF), which combines surface and distance data to incrementally create a consistent and scalable scene reconstruction. \\
Given the sequentially built G-GPDF model and the extracted 3D surface, we encode the scene into a 2D or 3D 360-degree raster-based projection. With this representation, pre-recorded binaural room impulse response filters are used to create the rendered spatial sound.
%First, we squash the 3D surface mesh points into their respective 2D coordinates, and then, these are used to create a 2D circular structure. We also do a simplified 3D cylindrical rasterisation while retaining the full 3D spatial structure. While the 2D circular rasterisation simplifies the 3D environment by projecting the mesh points onto a single 2D circle, the 3D cylindrical rasterisation represents both the horizontal and vertical points on the surface, capturing not only the azimuthal angles but also the height (elevation) of each point.  
%To show the potential of the representation to sonify a 3D environment, we develop various strategies for spatial audio feedback, facilitating intuitive sonification with auditory cues.

\subsection{2D Circular Representation}
Spatial audio feedback requires accurate but simple information, thus we propose to represent the incrementally built dense 3D map of the environment as a 2D circular sensor-centric representation. To ensure our representation aligns with human navigation requirements, we segment the ground and non-ground points. Any point above the ground and in the height range of a human that can potentially be an obstacle is classified as a non-ground point. Then, we project all these 3D non-ground points onto a 2D plane. 
% It is parallel to the camera's height. 
On this plane, every 3D point is reduced to its corresponding 2D coordinates, maintaining spatial relationships within the scene.

A circular grid centred at the location of the depth sensor is created by rasterising these 2D points. Rasterisation is faster and more efficient than raycasting, making it ideal for real-time processes and handling complex scenes. We do this by calculating the radial distance and azimuthal angle for every point with respect to the sensor. The azimuthal angle is derived by calculating the cross-product between the sensor's directional vector and the vector extending from the sensor to the point of interest. 

After that, this angle is normalised to fit inside a 360° circular representation. Each degree denotes a distinct angle surrounding the position of the sensor. The distance between the point and the sensor is used to calculate the radial distance. This enables us to rasterise each point, according to its distance from the sensor, onto a circular plane. 

This process of rasterisation yields a complete 360° depiction of the scene, with every point corresponding to a distinct angle and distance from the sensor. This ensures capturing the closest visible points surrounding the sensor's position. 
Sonifying the three-dimensional scene is especially advantageous with this minimalist circular representation. %We create a structure that maps to auditory specialisation by converting the spatial data into a 360-degree format with accurate azimuthal angles and radial distances. 
As we explain in the following section, each point in the circle can be associated with a specific sound, allowing for intuitive sound cues that represent the 3D environment. We ensure that the most pertinent spatial information is preserved by keeping only the closest points at each angle, which improves the audio feedback's clarity.

% The process begins by projecting the 3D mesh onto a 2D plane at the height of the camera. We then rasterise these 2D points onto a circle. This is done based on the radial distance and azimuth angle of the points with respect to the camera position. The azimuthal angle of each point is calculated by taking the cross-product of the directional vector of the camera and the vector of the point. This is further normalised to fit inside a 360° circular frame. The radial distance is the Euclidean distance from the camera to the point. This filters in only the points that are closest to the camera at each angle.

\subsection{3D Cylindrical Representation}
We take our representation further onto generating a 3D cylindrical structure instead of only a circle. This approach is guided by the intuition that the human is positioned within the cylinder. The reconstructed 3D mesh is rasterised onto a cylindrical surface to create the 3D cylindrical representation. This representation provides a structured, sensor-centric view of the surroundings. The cylindrical grid is discretised into vertical (elevation) and azimuthal intervals. The elevation spans heights between 0.1 and 2 meters above the sensor, while the azimuth is divided into 360-degree intervals.

Each 3D point in the scene is mapped to a specific azimuth-elevation pair, and only the closest point (in terms of radial distance) for each pair is retained. This process effectively reprojects the 3D points onto a simplified 2D representation of the scene. The reprojection occurs by collapsing the radial dimension, projecting all points onto a central 2D plane or line located at the center of the cylinder. The resulting 2D map provides a compact representation of the scene's geometry, where each point in the azimuth-elevation grid corresponds to the nearest surface point along the cylinder’s radial direction.

\subsection{Sonification}
%\cj{We can shorten this as needed}
We demonstrate the ability of our proposed representation to provide spatial information for cognitive sonification. An auditory framework is applied with the spatial representation to generate binaural auditory signals. A foundational aspect of the sonification process is to leverage the natural auditory perspective provided by human spatial hearing. In other words, humans naturally perceive the location of sounds around them and we design the representation to take advantage of spatial hearing perception. 

\begin{figure}[ht]
  \centering
  {\includegraphics[width=1.0\linewidth]{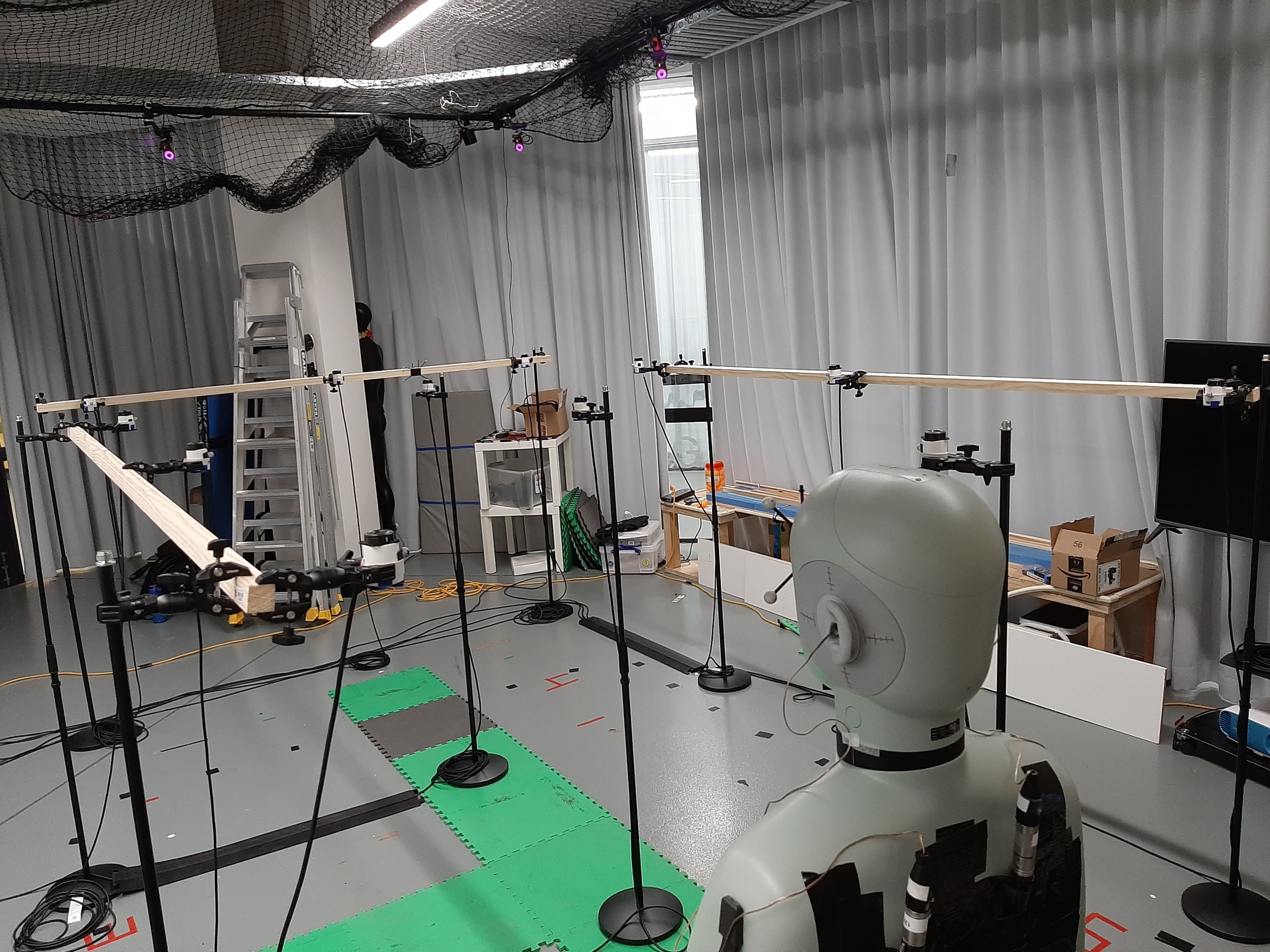}}
  \caption{The sound environment used for the BRIR recordings is shown. The HATS manikin is at the lower right.}
  \label{fig:brir_recording}
  %\vspace{-3ex}
\end{figure}
In this work, augmented or virtual spatial audio is enabled via the use of binaural room impulse responses (BRIRs) which are filters that mathematically specify the transformation of sound from a particular source location in a room to the ears of the individual listening to the map. There is a separate BRIR filter for each ear and each source location. Note that BRIR filters are unique for each individual in a perceptually relevant way because of morphological differences in ear shape. The data accumulated from the map representation has the spatial geometry of the room or environment. This information is sufficient to compute BRIRs using room simulation techniques. However, in this work, we use custom BRIRs that were prerecorded in a representative laboratory environment that includes a polypropylene carpet, a motion capture metallic frame, wooden beams with speakers mounted on them and other furniture (see Fig.~\ref{fig:brir_recording}). %Significantly, we restrict the source locations to near-field and far-field locations on the audiovisual horizon because source locations on the audiovisual horizon are perceptually more robust in augmented reality applications using generic, i.e., non-individualized, BRIRs.  

\begin{figure}[hbt]
  \centering
  \resizebox{0.65\linewidth}{!}{
  \subfloat{\includegraphics[height=4cm]{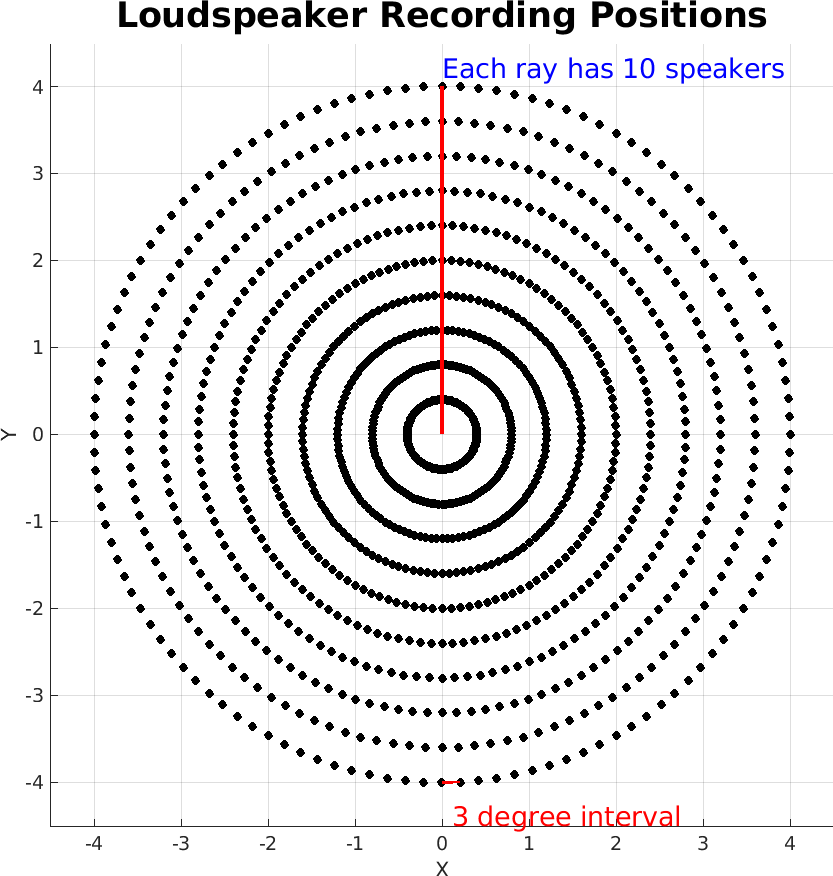}}}
  \caption{The efficiency performance of our proposed representation for spatial sonification with respect to different voxel resolutions.}
  \label{recording_positions}
  %\vspace{-3ex}
\end{figure}

The BRIR measurements were recorded using the Br\"{u}el \& Kj{\ae}r type 4128C Head and Torso Simulator (HATS) with two in-ear microphones together with a Br\"{u}el \& Kj{\ae}r type 9640 turntable. We arranged 10 small loudspeakers in a row at a height consistent with the audiovisual horizon of the HATS model. The 10 loudspeakers were positioned directly in front of the HATS model at zero degree of azimuth at distances varying from 0.4~m to 4~m in steps of 0.4~m. Using the turntable, the HATS model was rotated and BRIRs were recorded every $3^\circ$ of azimuth from $-177^{\circ}$ to $180^{\circ}$. %We also measured a sensor-to-ear transfer function for the Razer Anzu smart glasses and a headphone-to-ear transfer function for the Sennheiser HD 600 open headphones using the HATS model. Appropriate inverse filters were computed for both the smart glasses and headphones for playback of simulated binaural spatial audio signals. 
The BRIR files were saved as Spatially Oriented Format for Acoustics (SOFA) files~\cite{AES_2022}. The full loudspeaker recording positions are shown in Fig.~\ref{recording_positions}.

In order to demonstrate the suitability of the mapping representation for spatial sonification, we focus on the 2D circular representation. %We demonstrate two sonification modes of operation and various filtering operations. 
%One sonification mode, referred to as {\em circular ranging}, 
We sonify the distance to the surface for all directions around the circle counter-clockwise. It simulates a scanning ``length-changing cane". The cane automatically taps the closest surface in each discrete direction as it scans. As we mentioned, the BRIRs were recorded every $3^{\circ}$. Note that, our circular and cylindrical representations have $1^{\circ}$ resolution. To avoid sensory overload, we choose $10^{\circ}$ as the resolution to separate the 360-degree range data into 36 angular sectors. 

For each angular sector, the angle and range corresponding to the smallest range value are selected. The selected range data for each sector is then converted to binaural spatial audio by applying the appropriate BRIR filter to a `tap' sound. %chosen from the website freesound.org. %Ideally, the `tap' sound would relate to material properties, but we leave this for future work and use a constant `tap' sound in this work. 
Note that if a particular angular sector has not yet been scanned, a `woosh' sound is played instead of a `tap' sound. The distances encompassing the range data are scaled into the 0.4~m to 4~m range of the BRIRs using a constant scaling chosen appropriately for the given environment. Distances beyond 4~m are clamped to 4~m in the BRIRs range. This sonification distance range fits well with the perceptual properties of human distance perception, which is significantly poorer than human auditory direction perception~\cite{kolarik_auditory_2016} and we generally perceive distance in the near-field ($<$ 2~meters) better than the far-field ($>$ 2~meters). To further enhance the distance perception we apply a slight pitch shift to the tap sound. `Tap' sounds closer than 1.5~m are shifted down in pitch by four semitones and `tap' sounds further than 2.5~m are shifted up in pitch by four semitones. Our accompanying video shows these effects. Listening through headphones is recommended to hear the sounds at the correct azimuth location. %The selection of these parameters are arbitrary and simply chosen for demonstration purposes. %For the {\em circular ranging of objects} mode, we define a range discontinuity as a change of range, $\Delta$, with $\Delta > 0.5$~meters/degree. We also limit the number of objects in a single circular scan to the 5 nearest objects. Again, these parameters are arbitrary and chosen for demonstration purposes. %Interestingly, the range data from the 2D circular representation is discontinuous. There are jumps at the boundary between different objects because objects are separated and their distances vary randomly. We can use this observation advantageously to create an object-based sonification mode. In other words, we can segment the 2D circular range data at the discontinuities forming `pseudo objects' with an angular width defined by the discontinuities. Each object is then sonified only once in the direction and distance of its nearest point as the imaginary cane swings in a circle. We refer to this sonification mode as {\em circular ranging of objects}. 

%For both sonification modes, various filtering operations can be easily applied, providing `sensorisensor' feedback. In other words, the user controls a fundamental aspect of the sensor system -- the filtering operation -- and receives sensory feedback in the form of changes in the binaural auditory signal. We now describe several filtering operations:
% \begin{itemize}
%     \item {\em field-of-view sonification filter:} during the circular operation, sounds are only generated for specified ranges of azimuthal angle.
%     \item {\em distance sonification filter:} during the circular operation, sounds are only generated for specified ranges of distance.
%     \item {\em object counting filter:} during the ``circular ranging of objects'', sounds are only generated for a limited number of objects, controlling object clutter.
% \end{itemize}

%The details for the demonstration of the two sonification modes are now described. 

\section{Evaluation}
To the best of our knowledge, no open-source state-of-the-art spatial sonification frameworks are available for comparison. Therefore, we quantitatively evaluate the proposed representation for spatial sonification in terms of a) efficiency, b) representation accuracy and c) coverage with respect to using depth images, d) qualitatively demonstrate the representation performance with a dynamic object and e) suitability for sonification with respect to Euclidean distance fields. Our framework is implemented in C++ based on ROS1. All experiments were run on 12th Gen Intel® Core™ i5-1245U with 12 cores.

The first dataset is \emph{the Cow and Lady}\cite{Voxblox}, which includes fibreglass models of a large cow and a lady standing side by side in a room.
The dataset consists of RGB-D point clouds collected using a Kinect 1 camera, with sensor trajectory captured using a Vicon motion system. We use all frames of the dataset, which covers major information of the scene over 360 degrees. 
We use the ground truth map of \emph{the Cow and Lady}, which allows us to perform a proper quantitative evaluation for representation accuracy. As our second dataset, we run the online SLAM in a room with a live RealSense RGB-D camera. The room has a human as a dynamic object. The poses are computed from the VINS~\cite{vins_rgbd} localisation framework as described above. 
\begin{figure}[hbt]
  \centering
  \resizebox{0.8\linewidth}{!}{
  \subfloat{\includegraphics[height=4cm]{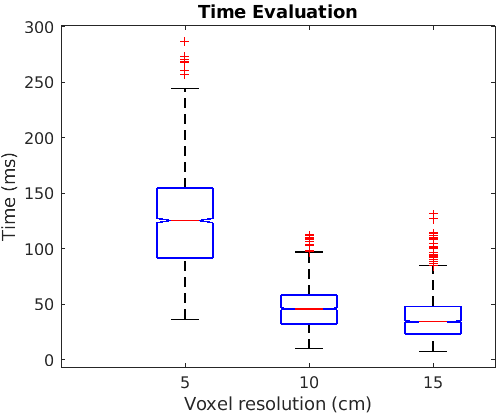}}}
  \caption{The efficiency performance of our proposed representation for spatial sonification with respect to different voxel resolutions.}
  \label{time}
  %\vspace{-3ex}
\end{figure}

\begin{figure*}[hb]
	\centering
    \resizebox{\linewidth}{!}{
	\subfloat[\label{rmse_5}]{\includegraphics[height=4.5cm]{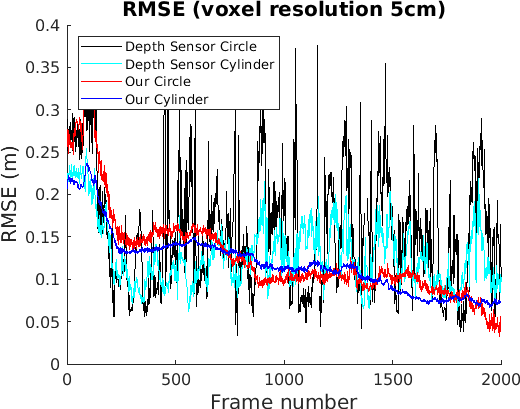}}
	%\resizebox{\linewidth }{!}{
    \subfloat[\label{rmse_10}]{\includegraphics[height=4.5cm]{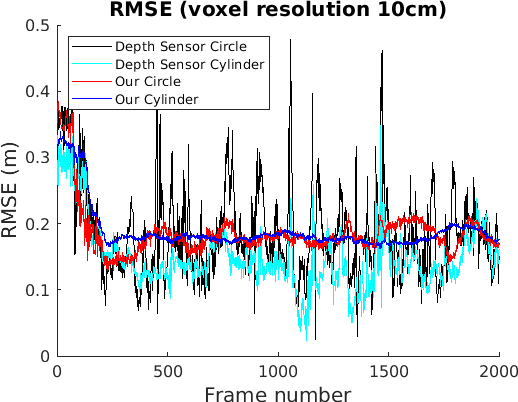}}
    %\resizebox{\linewidth }{!}{
	\subfloat[\label{rmse_15}]{\includegraphics[height=4.5cm]{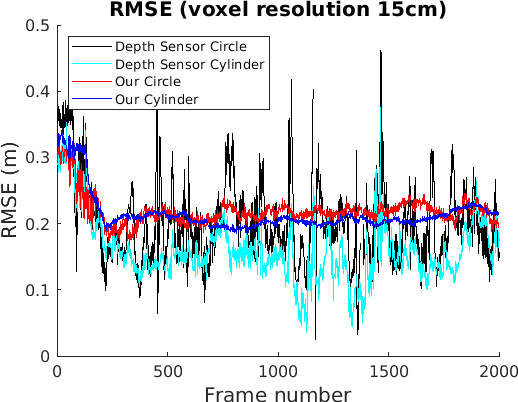}}
    }
	\caption{Quantitative comparisons of the accuracy in RMSE on \emph{the cow and lady} dataset with varying voxel resolutions.}
	\label{rmse_2d_3d}
\end{figure*}
\subsection{Time Evaluation}
We first evaluate the time consumption to show our representation can provide online performance for sonification. Fig.~\ref{time} shows the statistical computational time for \emph{the Cow and Lady} dataset. Note that our framework has the free-space carving methods~\cite{wu2024vdb} enabled to update the moving objects in the scene. 
For the sake of simplicity, we compute the time for all the processes, including fusion, 2D circle, and 3D cylinder. It illustrates that our representation is efficient with varying resolutions for online performance.

\subsection{Representation Accuracy}
For the accuracy evaluation, we use the raw depth images in the same way as in our framework to compute the 2D circle and 3D cylinder. Note that we apply the ground truth point cloud of \emph{the Cow and Lady} to calculate the ground truth circle and cylinder. We then quantitatively compare our circle and cylinder representations and depth images with different map resolutions against the ground truth. 
As demonstrated in Fig.~\ref{rmse_2d_3d}, at the beginning of the dataset, all RMSEs are high due to noisy measurements when the drone is taking off. Then the RMSE drops due to more information is collecting along the steady exploration. 
The RMSE of the depth sensor circle and cylinder is noisy due to the lack of fusion over time. Our accuracy outperforms depth representations in different resolutions.
Note that occasionally, the depth sensor RMSE has lower values than ours, and this is because the depth sensor its only calculated in the field of view of the camera. Therefore, compared to GT and ours, only a few points in the depth circle and cylinder have information and are being evaluated for RMSE.
% In Fig.~\ref{teaser}, we qualitatively show our mesh coloured by fused LiDAR intensity and the distance field output. Fig.~\ref{teaser1} shows the incrementally built dense reconstruction. We zoomed in Fig.~\ref{teaser2} to show the textures on the ceiling inside the corridor and Fig.~\ref{teaser3} the stairs and windows at the corner of the quad. 

% \begin{figure}[ht]
%   \centering
%   \resizebox{\linewidth }{!}{
%   \subfloat[ \label{teaser1}]{\includegraphics[height=4cm]{figures/RMSE_circel_cylinder.png}}}
%   \caption{Our proposed representation for spatial sonification. a)  b) }
%   \label{teaser}
%   \vspace{-3ex}
% \end{figure}
% four figures in total
% add another figure for depth

\begin{figure*}[ht]
	\centering
    \resizebox{1.0\linewidth }{!}{
	\subfloat[]{\includegraphics[height=4.5cm]{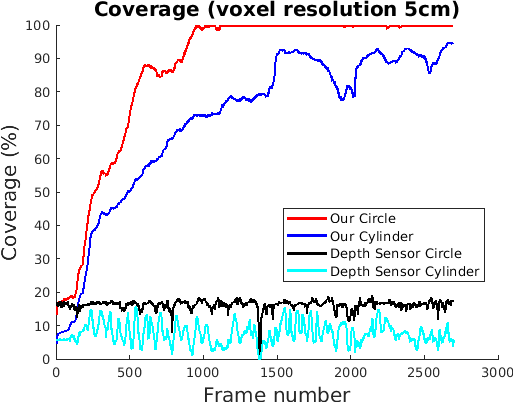}}
	%\resizebox{\linewidth }{!}{
    \subfloat[]{\includegraphics[height=4.5cm]{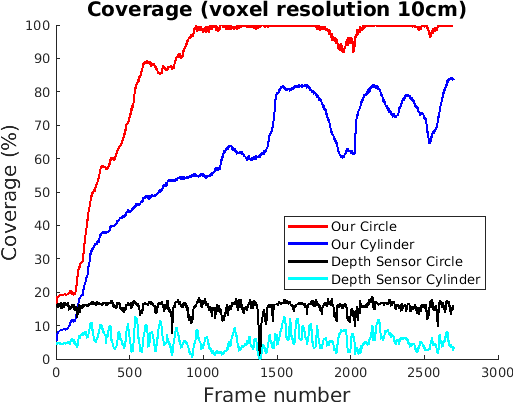}}
	%\resizebox{\linewidth }{!}{
    \subfloat[]{\includegraphics[height=4.5cm]{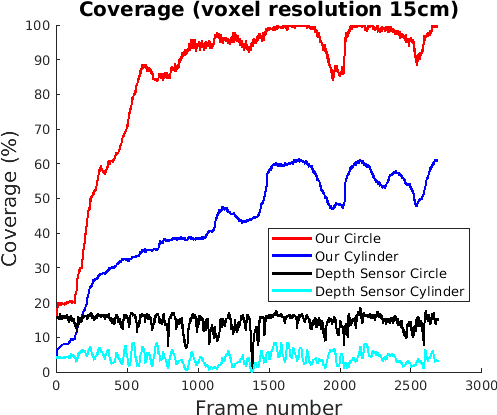}}
    }
    
	\caption{Quantitative comparisons of the coverage on the cow and lady dataset with varying voxel resolutions.}
	\label{coverage_2d_3d}
    \vspace{-2ex}
\end{figure*}

\subsection{Coverage Over Frames}
We compare the coverage of our framework against the depth sensor again. Due to the ground truth having missing areas, we use the fully reconstructed mesh as the benchmark to compute the coverage. Demonstrating in Fig.~\ref{coverage_2d_3d}, as the mapping voxel resolution varies from 5cm to 15cm, the coverage of the depth representations remains the same. The depth circle covers below 20\%, and the cylinder covers around 10\%. Our incremental representation accumulates and fuses measurements as the coverage grows over frames. Note that the coverage gets up and down occasionally due to passing through unexplored areas to gather more information. With a high resolution of 5cm, our circle reaches 100\% coverage, and our cylinder covers over 90\% at the end.

\begin{figure*}[t]
	\centering
    \resizebox{\linewidth }{!}{
	\subfloat[\label{dynamic_1}]{\includegraphics[height=5cm]{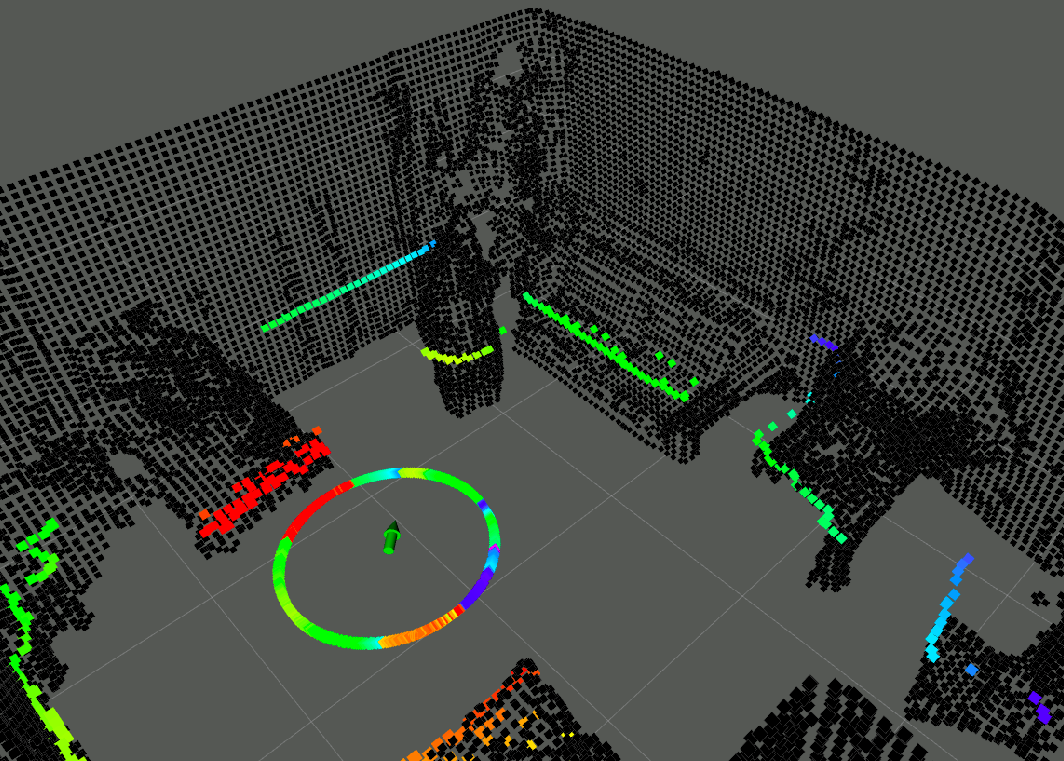}}
 	\subfloat[\label{dynamic_2}]{\includegraphics[height=5cm]{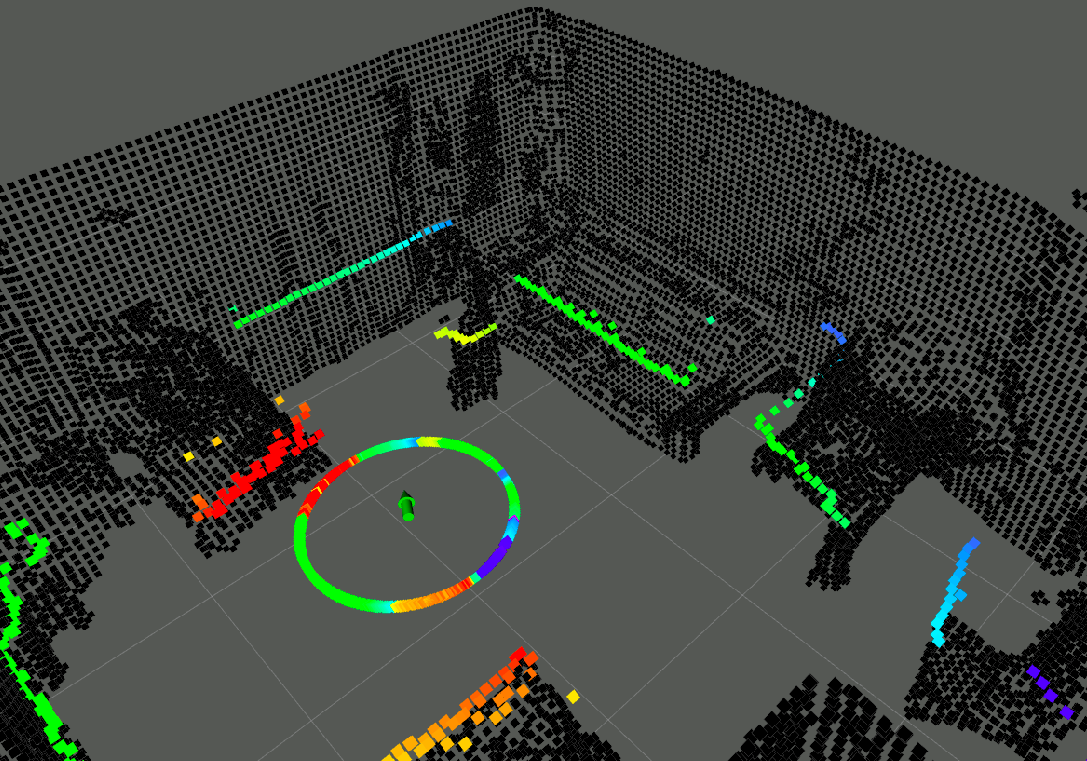}}
	%\resizebox{\linewidth }{!}{
    \subfloat[\label{dynamic_3}]{\includegraphics[height=5cm]{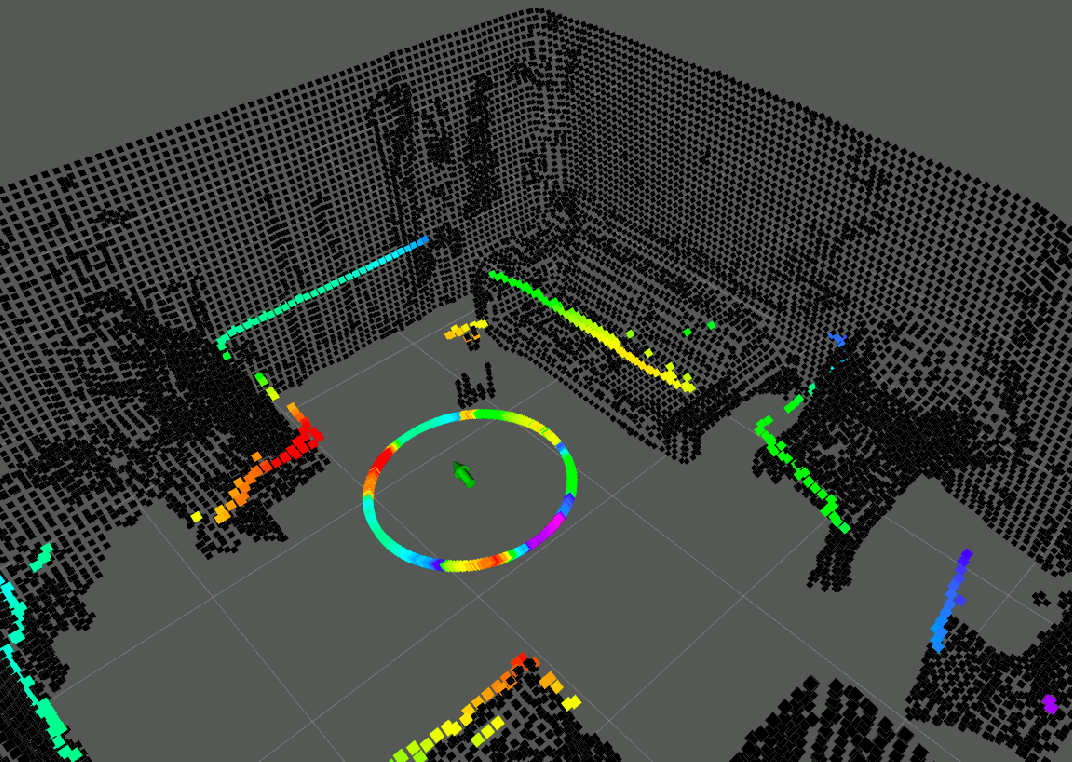}}
	%\resizebox{\linewidth }{!}{
    \subfloat[\label{dynamic_4}]{\includegraphics[height=5cm]{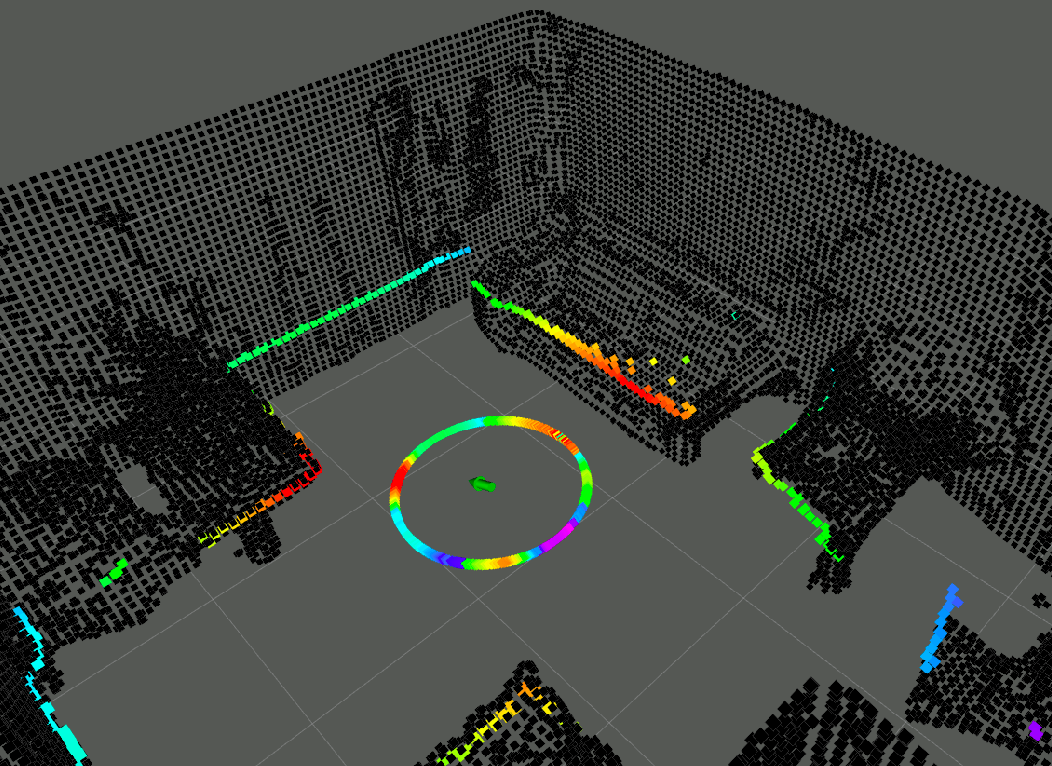}}
    }
	\caption{We map the scene with a live camera online to show the ability to deal with dynamic objects.}
	\label{dynamic_circle}
    \vspace{-3ex}
\end{figure*}
\subsection{Dynamic Moving Object Validation}
We aim to show that our representation has the ability to deal with dynamic objects, which is very common in the real-world scenario. Our framework takes poses and points as raw input, so we use localisation frameworks such as VINS~\cite{vins_rgbd} to provide proper poses. We move a live camera to map the scene online. In Fig.~\ref{dynamic_circle}, our 2D circle is fully reconstructed in 360 degrees, and we can visual object projection on the circle for each angle clearly. From Fig.~\ref{dynamic_1} to Fig.~\ref{dynamic_4}, there was an object in the scene and later moved away. Our circle is updated with respect to the latest status of the scene. 

\begin{figure}[ht]
  \centering
  \resizebox{1.0\linewidth}{!}{
  \subfloat[\label{arrow_ours}]{\includegraphics[height=4cm]{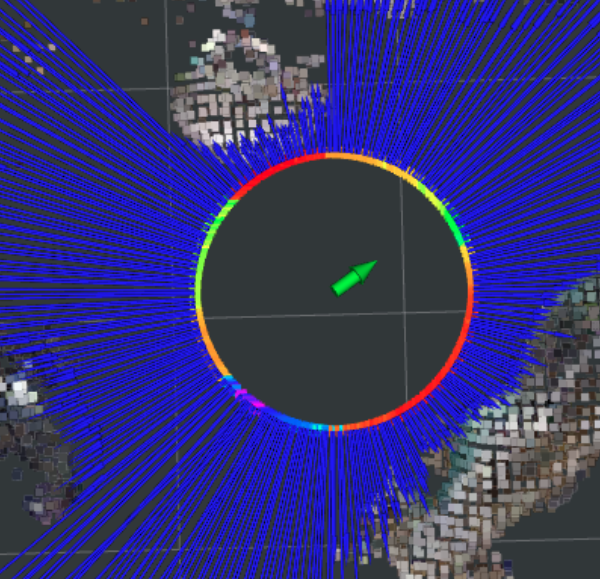}}
  \subfloat[\label{arrow_esdf}]{\includegraphics[height=4cm]{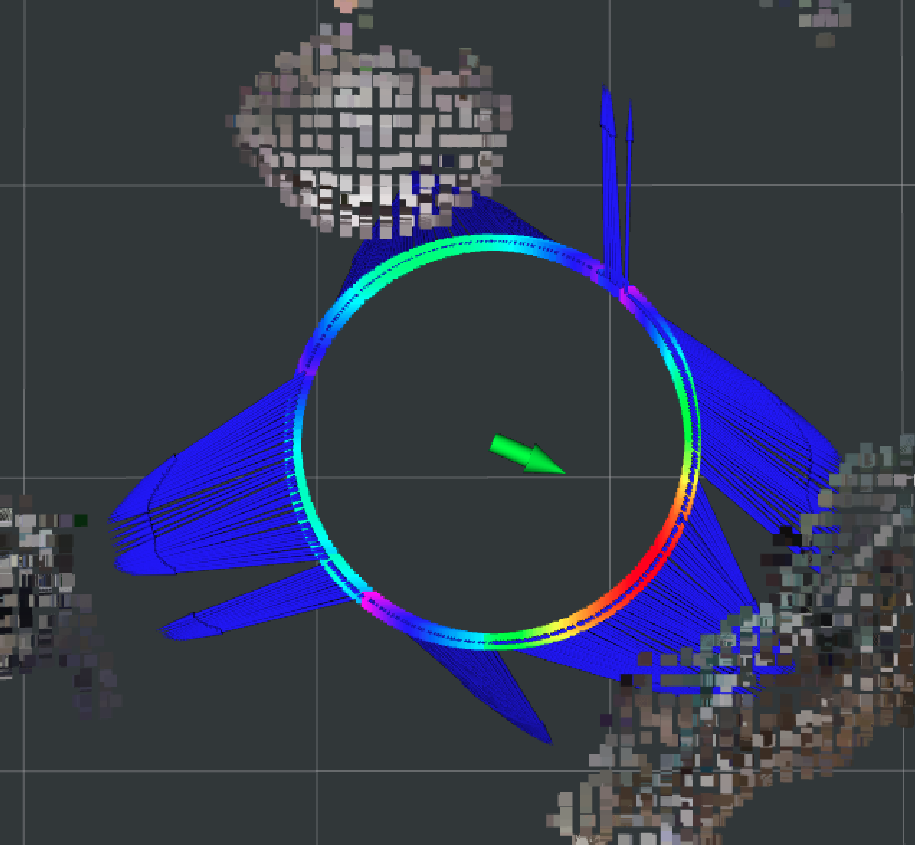}}
  }
  \caption{a) Our circular representation encodes the radial distance at a given angle over the scan. b) EDF, however, encodes the distance and directions to the closest surfaces.}
  \label{compare_esdf}
  \vspace{-3ex}
\end{figure}
\subsection{Compare to EDF}
This section aims to show the suitability of our representation for sonification over the Euclidean distance fields (EDF), which is better suited for path planning and navigation. For the EDF~\cite{wu2024vdb}, it usually has a distance value and the gradient of the distance given the point's location in the space. It always measures the distance and direction from the point to the closest surface. As we show in Fig.~\ref{arrow_ours}, our circle has bearing angles pointing 360 degrees, and for each bearing, we have the closest distance along the ray (the ray arrow is shown in blue). However, as we see from Fig.~\ref{arrow_esdf}, we query the EDF for each point on the circle, and this gives us the distance and gradient to the closest surfaces. The gradients of distances (shown in blue) on the circle slip around as the closest surface could be in any direction. This is not suitable for sonification as the bearing does not vary sequentially. 

\begin{figure*}[t]
	\centering
    \resizebox{\linewidth }{!}{
	\subfloat[\label{dynamic_1}]{\includegraphics[height=5cm]{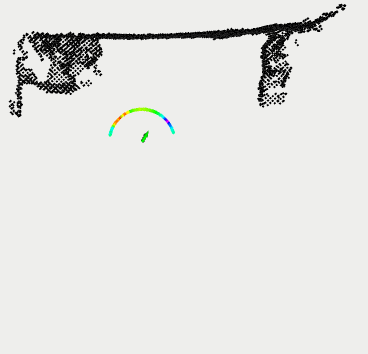}}
 	\subfloat[\label{dynamic_2}]{\includegraphics[height=5cm]{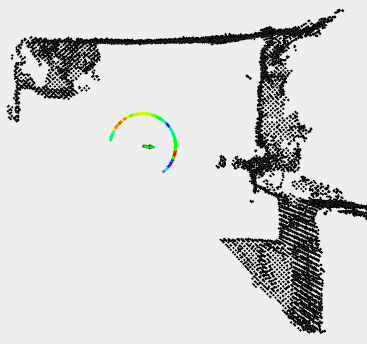}}
	%\resizebox{\linewidth }{!}{
    \subfloat[\label{dynamic_3}]{\includegraphics[height=5cm]{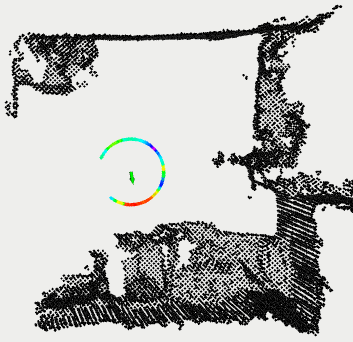}}
	%\resizebox{\linewidth }{!}{
    \subfloat[\label{dynamic_4}]{\includegraphics[height=5cm]{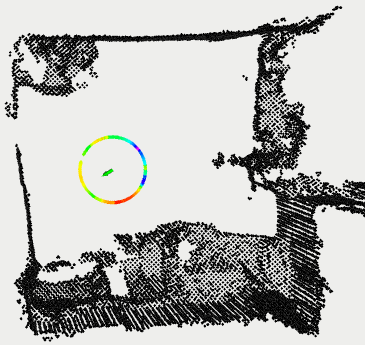}}
    }
	\caption{Incrementally built scene with a live depth camera following our framework mapping, self-localising, and maintaining the proposed sonification representation.}
	\label{online_demonstration}
    \vspace{-3ex}
\end{figure*}

\subsection{Sonification Demonstration}
In our accompanying video and Fig.~\ref{online_demonstration}, we show an incrementally built and sonified room-sized environment using the proposed framework. Incremental localisation (using VINS~\cite{vins_rgbd}), mapping (with VDB-GPDF~\cite{wu2024vdb}), circular rasterization and sonification are shown in real-time. Listening through headphones is recommended to hear the sounds at the correct azimuth location.

\section{Conclusion}
We propose an online, incrementally built representation to provide spatial auditory information of unknown environments. This approach has the potential to aid individuals with visual impairments. By leveraging a sensor-centric mapping structure based on depth sensors, we effectively convert 3D environments into 360-degree auditory representations. Our approach, which includes the VDB-GPDF, plane projection, rasterisation and binaural room impulse responses, ensures fast accurate, complete and perceptually robust mapping for sonification. We demonstrate its performance in both static and dynamic environments. The successful implementation and evaluation of this framework highlight its potential to significantly improve the quality of life for visually impaired individuals by providing an intuitive and efficient means of navigating their surroundings. In future work, we aim to perform user studies targeting to reduce the cognitive load of the users for the 3D representation.

\begin{acronym}[AAAAAA]
    
    \acro{UTS}{University of Technology Sydney }
    \acro{RI}{Robotics Institute }
    \acro{FEIT}{Faculty of Engineering and Information Technology}
    \acro{WiEIT}{Women in Engineering and IT}
    \acro{1D}{One-Dimensional }
    \acro{2D}{Two-Dimensional }
    \acro{2.5D}{Two-and-a-Half-Dimensional}
    \acro{3D}{Three-Dimensional }
    \acro{GP}{Gaussian Process }
    \acro{GPIS}{Gaussian Process Implicit Surfaces }
    \acro{MAP}{Maximum A Posteriori }
    \acro{MLE}{Maximum Likelihood Estimation }
    \acro{ICP}{Iterative Closest Point }
    \acro{MVE}{Multi-View Environment }
    \acro{OG}{Occupancy Grid }
    \acro{CHOMP}{Covariant Hamiltonian Optimization for Motion Planning}
    \acro{FIESTA}{Fast Incremental
Euclidean DiSTAnce Fields}
    \acro{GPUs}{Graphics Processing Units }
    \acro{KD}{K-Dimensional}
    
    \acro{Log-GPIS}{Log-Gaussian Process Implicit Surfaces }
    \acro{LiDAR}{Light Detection And Ranging Sensor }
    \acro{SLAM}{Simultaneous Localisation and Mapping }
    \acro{TDF}{Truncated Distance Field }
    \acro{EDF}{Euclidean Distance Field }
    \acro{SDF}{Signed Distance Field }
    \acro{TSDF}{Truncated Signed Distance Field }
    \acro{ESDF}{Euclidean Signed Distance Field }
    
    \acro{GPIS-SDF}{GPIS with signed distance function }
    
    \acro{RGB}{Red-Green-Blue }
    \acro{RGB-D}{Red-Green-Blue-Depth }
    \acro{RMSE}{Root Mean Squared Error }
    
    %\acro{SE(3)}{Special Euclidean group in three dimensions }
%    \acro{GMM}{Gaussian Mixture Models }
%    \acro{NDT}{Normal Distribution Transform }
%    \acro{GPLVM}{Gaussian Process Latent Variable Models }
    \acro{CI}{Conditional Independent }
    \acro{FP}{Forward Propagation}
    \acro{BP}{Backward Propagation}
    \acro{D-SKI}{Structured Kernel Interpolation framework with Derivatives}
    \acro{SKI}{Structured Kernel Interpolation method}
    \acro{D-SKI-CI-Fusion}{Structured Kernel Interpolation with Derivatives and Conditional Independent Fusion}
    \acro{MVMs}{Matrix-Vector Multiplications method}

    \acro{PDE}{Partial Differential Equation }
    \acro{Log-GPIS-MOP}{Log-Gaussian Process Implicit Surface for Mapping, Odometry and Planning }
    \acro{RANSAC}{Random Sample Consensus}
    \acro{Dynamic-GPDF}{Dynamic Gaussian Process Distance Fields}
\end{acronym}

\bibliographystyle{IEEEtran}
\bibliography{reference}

\end{document}